\documentclass{SSBAtran}

\usepackage[T1]{fontenc}
\usepackage{graphicx}
\usepackage{xcolor}
\usepackage{makecell}
\usepackage{amsmath}
\usepackage{float}

\begin{document}
\title{Comparing Next-Day Wildfire Predictability of MODIS and VIIRS Satellite Data}
\author{\IEEEauthorblockN{%
        Justus Karlsson\IEEEauthorrefmark{2},
        Yonghao Xu\IEEEauthorrefmark{2}\IEEEauthorrefmark{1},
        Amanda Berg\IEEEauthorrefmark{2}\IEEEauthorrefmark{3} and
        Leif Haglund\IEEEauthorrefmark{2}\IEEEauthorrefmark{3}}
    \IEEEauthorblockA{\IEEEauthorrefmark{2}Computer Vision Laboratory, Linköping University, Sweden}
    \IEEEauthorblockA{\IEEEauthorrefmark{3}Maxar Intelligence, Linköping, Sweden}
    {\tt\{\small justus.karlsson, yonghao.xu, amanda.berg, leif.haglund\}@liu.se
    }
    \IEEEauthorblockA{\IEEEauthorrefmark{1}Corresponding author}
}

\maketitle              
\begin{abstract}
    Multiple studies have performed next-day fire prediction using satellite imagery. Two main satellites are used to detect wildfires: MODIS and VIIRS. Both satellites provide fire mask products, called MOD14 and VNP14, respectively. Studies have used one or the other, but there has been no comparison between them to determine which might be more suitable for next-day fire prediction.

    In this paper, we first evaluate how well VIIRS and MODIS data can be used to forecast wildfire spread one day ahead. We find that the model using VIIRS as input and VNP14 as target achieves the best results. Interestingly, the model using MODIS as input and VNP14 as target performs significantly better than using VNP14 as input and MOD14 as target. Next, we discuss why MOD14 might be harder to use for predicting next-day fires. We find that the MOD14 fire mask is highly stochastic and does not correlate with reasonable fire spread patterns. This is detrimental for machine learning tasks, as the model learns irrational patterns. Therefore, we conclude that MOD14 is unsuitable for next-day fire prediction and that VNP14 is a much better option. However, using MODIS input and VNP14 as target, we achieve a significant improvement in predictability. This indicates that an improved fire detection model is possible for MODIS. The full code and dataset is available online.

\end{abstract}
\section{Introduction}

Wildfires are getting more frequent and severe due to climate change \cite{wildfire-season,Westerling2006}. Being able to more accurately predict wildfire behavior can help with wildfire management and response \cite{Thompson2016,Thompson2019}. One option for next-day fire prediction is to use satellite imagery. MODIS and VIIRS \cite{mod14,vnp14} satellites provide twice-daily global coverage and are equipped with both infrared and visible sensors suitable for detecting wildfires. These factors make satellite imagery a very powerful tool for wildfire detection and prediction, compared with, for example, aerial imagery, since the coverage is consistent both spatially and temporally. Furthermore, the wealth of data makes machine learning a good candidate for learning wildfire behavior.

Both MODIS and VIIRS provide products for active fire detection, called MOD14 \cite{mod14} and VNP14 \cite{vnp14}, respectively. VNP14 is based on the same contextual algorithm for detecting fires as MOD14. Several studies have performed next-day fire prediction using either MOD14 \cite{mod14-3,mod14-1,mod14-4,mod14-2} or VNP14 \cite{vnp14-1,vnp14-4,vnp14-2,vnp14-3}. However, there has not yet been a comparison between the two data sources for the task of next-day fire prediction. Having a better understanding of the pros and cons of the two fire mask products would be very beneficial for future research on wildfire detection.

\begin{figure}[htbp]
    \centering
    \includegraphics[width=0.45\textwidth]{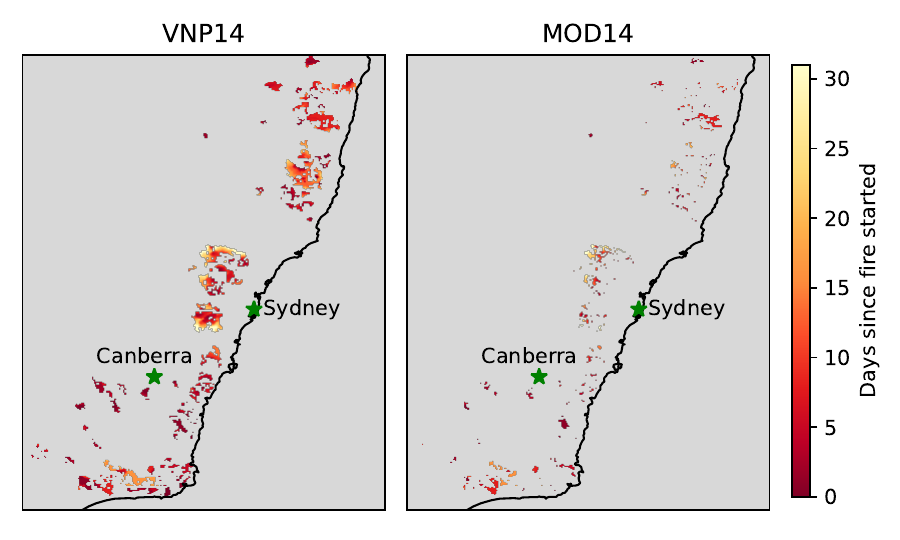}
    \caption{Fire progression during the wildfire period 2019-10-01 to 2020-01-31 according to VNP14 (left) and MOD14 (right). The color scale shows progression of each individual fire measured from the ignition date of said fire. The area shown, southeast Australia, is a subsection of the study area. Each pixel in the image represents about 1 km.}
    \label{fig:south-east-australia}
\end{figure}

In this paper, we conduct a two-step study to assess which data source is better suited for next-day fire prediction. First, we compare the predictability of VIIRS and MODIS for next-day fire prediction. To restrict the amount of data for this study, we only gather data from mainland Australia. Data from two time periods are collected.
For the training and validation set, the period from 2019-10-01 to 2020-01-31 was used. For the test set, the period from 2023-10-01 to 2024-01-31 was used. Both these periods contained widespread wildfires in Australia. A visualization of fires from the training and validation set is shown in Figure \ref{fig:south-east-australia}. We frame the task as a binary-classification image segmentation problem. The comparison is done by training identical deep learning models, where the only difference is the input and target data. Predictability is not conclusive evidence for real-world accuracy. For example, a trivial case where the product only outputs zeros is 100\% predictable, but not very useful. Therefore, the second step is to assess why one of them might be easier for a deep learning model to predict. The goal of such an assessment is to reach a conclusion on which data source leads to next-day fire prediction that correlates best with real-world fire spread. To facilitate further research and evaluation on MODIS and VIIRS for next-day fire prediction, the code and data are made public\footnote{https://github.com/justuskarlsson/wildfire-mod14-vnp14}. To summarize, the main contributions of this paper are:
\begin{enumerate}
    \item A quantitative analysis of the predictability of MOD14 and VNP14 for next-day fire prediction using deep learning models.
    \item A qualitative analysis of the fire spread behavior according to MOD14 and VNP14.
    \item A public dataset and codebase enabling further research comparing MODIS and VIIRS for wildfire prediction.
    \item A framework for constructing a global dataset for next-day fire prediction. Even though this study is restricted to Australia, the code and data retrieval is compatible with all regions globally.
\end{enumerate}

\section{Related work}

First, we describe the two fire mask products, MOD14 and VNP14. Then we discuss previous work using MOD14 and VNP14 for next-day fire prediction.


\subsection{MOD14 and VNP14 fire mask products}
In 2006, Justice et al. \cite{mod14} described their MODIS fire products (including MOD14).
The 3.929--3.989\ \textmu m band at 1 km spatial resolution was the primary band used for fire detection. A number of other bands at different wavelengths and spatial resolutions were used to detect false alarms, such as sun glint.
To perform fire detection, a contextual algorithm was used that first rejected obvious non-fire pixels, then analyzed neighboring pixels to estimate background signal, and finally applied threshold tests comparing brightness temperatures to detect fires while eliminating false detections.

In 2013, Schroeder et al. \cite{vnp14} presented their VNP14 active fire product based on VIIRS data. The algorithm builds on the MOD14 algorithm by Justice et al. \cite{mod14}. A major difference compared with MOD14 is that the VNP14 fire detection only uses bands at a spatial resolution of 375 m. Schroeder et al. assert that this leads to ''improved consistency of fire perimeter delineation for biomass burning lasting multiple days,'' compared with MOD14.
The 3.550--3.930\ \textmu m band was the primary band used for fire detection in VNP14. The contextual algorithm used in VNP14 is similar to the one used in MOD14 but adapted to the different band characteristics of the VIIRS sensor.

In addition to differences in spatial resolution and algorithm design, the observation times of the platforms also differ, which can influence fire detection capabilities. Specifically, the Terra platform used for MOD14 captures images at 10:30 and 22:30 local time. The S-NPP/VIIRS platform, used for VNP14, captures images at 13:30 and 01:30 local time.

\subsection{Previous work using MOD14}
Huot et al. \cite{mod14-1} used Google Earth Engine (GEE) \cite{gee} to retrieve MOD14 fire masks and auxiliary data over the United States. They make no mention of VNP14, however, at the time of writing their publication, VNP14 was not yet available on GEE. They evaluate multiple deep learning models on the next-day fire prediction task and frame it as an image segmentation problem.
In a subsequent publication, Huot et al. \cite{mod14-3} used a similar approach, but again, they make no mention of potentially using VNP14. Their analysis was limited to data accessible through Google Earth Engine.
Prapas and Kondylatos et al. \cite{mod14-2} used MOD14 data together with historical burned areas from the European Forest Fire Information System (EFFIS). Their study is restricted to the Eastern Mediterranean region. In contrast with \cite{mod14-3,mod14-1}, they focused specifically on predicting fire \emph{ignition}. The problem was also set up as image classification as opposed to image segmentation, since only the pixel in the center of each image was evaluated. In a subsequent publication, Prapas and Kondylatos et al. \cite{mod14-4} used a similar approach as \cite{mod14-2}. VNP14 is not mentioned as a potential fire mask source here either.

In summary, the papers using MOD14 make no mention of VNP14.

\subsection{Previous work using VNP14}
Ali et al. \cite{vnp14-1} present a dataset for next-day fire spread prediction. The dataset uses VNP14 for their fire masks. The two main reasons they mention for using VNP14 instead of MOD14 are that VNP14 has a higher spatial resolution and that it has a higher number of fire detections. They make no direct evaluation of the quality between MOD14 and VNP14. Gerard et al. \cite{vnp14-2} present the dataset \emph{WildfireSpreadTS} for next-day fire spread prediction. No thorough discussion of the choice between MOD14 and VNP14 is made, but it seems that increased spatial resolution was the main reason for using VNP14. Zhao et al. \cite{vnp14-3} present a multi-task dataset called \emph{TS-SatFire}. They label one of the tasks as ''wildfire progression prediction'', which fits into our label of next-day fire prediction. The fire masks used for the next-day fire prediction tasks are from VNP14. The increased spatial resolution is cited as the reason for choosing VNP14 over MOD14. Ali et al. \cite{vnp14-4} present a dataset for next-day wildfire spread prediction. The fire masks used are from VNP14, but no discussion of the choice between MOD14 and VNP14 is made.

In summary, the papers using VNP14 primarily mention the increased spatial resolution as the reason for using VNP14 over MOD14.

\subsection{Comparison of MOD14 and VNP14}
We can see a mix of MOD14 and VNP14 in the literature, but there has not been a definitive comparison of which data source is better for predicting fire spread. The motivation for using one or the other is typically not well formulated. One stated argument for using VIIRS is the higher spatial resolution. This argument makes no judgment of the quality of the data, even though it might be natural to assume that higher spatial resolution correlates with better quality. Studies that use MODIS typically do not motivate its use in comparison with VIIRS. One reason might be data availability. In remote sensing, it is common to use Google Earth Engine (GEE) \cite{gee} for data retrieval. Although both VNP14 and MOD14 are available on GEE, VNP14 is only available from 2023-09-03, whereas MOD14 is available from 2000-02-24. Using NASA's own data portal (LP-DAAC), VNP14 can be retrieved back to 2012-01-17.

\section{Method}
First, the method of selecting, retrieving, and processing the data is described. We then list some choices and their motivations in relation to the predictability experiments. The underpinning motivation is to maximize the fairness of the comparison between VIIRS and MODIS. Finally, we briefly describe how the qualitative assessment was set up.

\subsection{Data retrieval, selection and processing}
We focus our study on Australia, a country that has experienced severe wildfires \cite{xu2024sen2fire}, while also limiting the volume of data to be retrieved. For the training and validation dataset, we use the period from 2019-10-01 to 2020-01-31. The test dataset is from 2023-10-01 to 2024-01-31. These two periods contained a large number of fire detections in both MOD14 and VNP14. The period from October to January is part of the eastern Australian wildfire season \cite{wildfire-season}, which also motivated the specific months used.

\begin{figure}[htbp]
    \centering
    \includegraphics[width=0.45\textwidth,trim=0 80 0 80,clip]{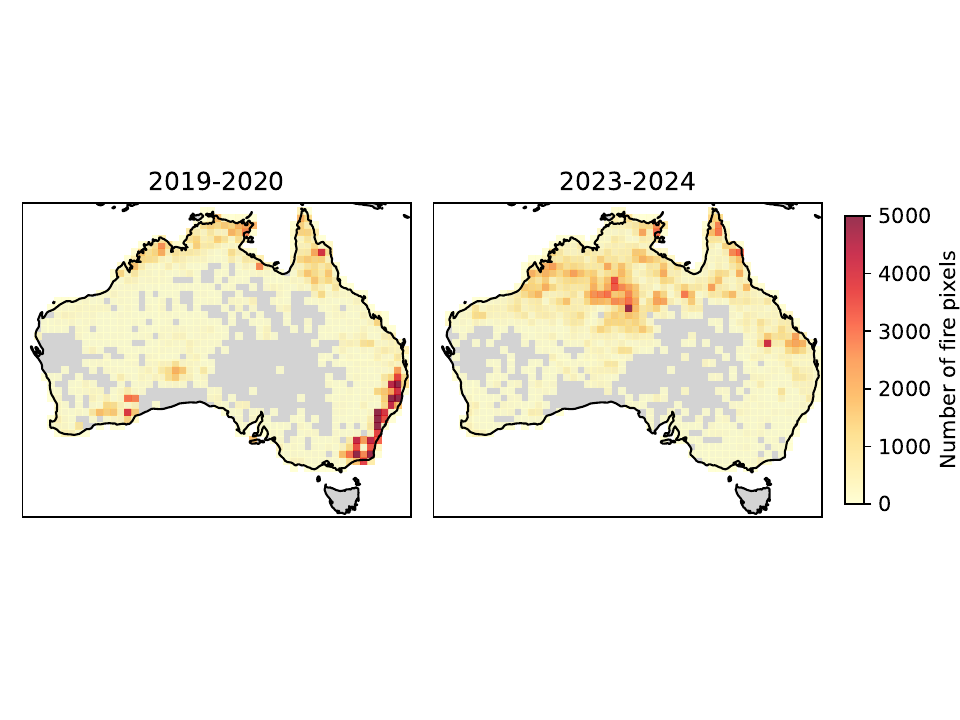}
    \caption{Fire detections (VNP14) in Australia during the two study periods. The left image shows fire detections from 2019-10-01 to 2020-01-31, which is used for training and validation. The right image shows fire detections from 2023-10-01 to 2024-01-31, which is used for testing. Both periods show significant wildfire activity, albeit in slightly different areas.}
    \label{fig:study-periods}
\end{figure}

The training and validation dataset is split uniformly based on x and y grid cells by 75\% and 25\% respectively. The test dataset is instead separated from the training/validation dataset temporally. This split results in 2,654 training samples, 753 validation samples, and 4,710 test samples.

As input data, we use the raw bands from MODIS and VIIRS. Furthermore, we also include weather and drought data from the input day. This data is from ERA5 \cite{era5} and KBDI \cite{drought} respectively. The raw bands data (Level 1 products) for MODIS and VIIRS is available from LAADS. The fire mask products are available from LP-DAAC. The Python package \textit{earthaccess} \cite{earthaccess} is used to gather data from both data portals. The package allows for retrieving images based on the desired date range and area of interest. The data products retrieved via earthaccess are shown in Table \ref{tab:data-products}. The weather\footnote{ECMWF/ERA5\_LAND/DAILY\_AGGR} and drought\footnote{UTOKYO/WTLAB/KBDI/v1} data are retrieved from Google Earth Engine (GEE). Similarly to \textit{earthaccess}, this service also allows for easy filtering according to the desired date range and area of interest.

\begin{table}[h]
    \centering
    \caption{List of data products retrieved via earthaccess. The products are divided by their use, sensor name, and source. LAADS and LP-DAAC are NASA data distribution services.}
    \begin{tabular}{l|l|l}
        \multicolumn{3}{c}{\textbf{Data Products}}                      \\
        \hline
        Use         & Product                             & Data portal \\
        \hline
        Geolocation & MOD03 \cite{mod03}                  & LAADS       \\
        Geolocation & VNP03IMG \cite{vnp2,vnp1,vnp4,vnp3} & LAADS       \\
        Geolocation & VNP03MOD \cite{vnp2,vnp1,vnp4,vnp3} & LAADS       \\
        1 km bands  & MOD021KM \cite{mod021km}            & LAADS       \\
        500 m bands & MOD02HKM \cite{mod02hkm}            & LAADS       \\
        250 m bands & MOD02QKM \cite{mod02qkm}            & LAADS       \\
        375 m bands & VNP02IMG \cite{vnp2,vnp1,vnp4,vnp3} & LAADS       \\
        750 m bands & VNP02MOD \cite{vnp2,vnp1,vnp4,vnp3} & LAADS       \\
        Fire Mask   & MOD14 \cite{mod14-product}          & LP-DAAC     \\
        Fire Mask   & VNP14IMG \cite{vnp14-product}       & LP-DAAC     \\
        \hline
    \end{tabular}
    \label{tab:data-products}
\end{table}

\begin{table}[h]
    \centering
    \caption{Overview of MODIS bands and their resolutions. The bands are divided into reflective and emissive types.}
    \begin{tabular}{l|l|l}
        \multicolumn{3}{c}{\textbf{MODIS Bands}}     \\
        \hline
        Resolution & Band               & Type       \\
        \hline
        250 m      & B1--B2             & Reflective \\
        500 m      & B3--B7             & Reflective \\
        1 km       & B8--B19, B26       & Reflective \\
        1 km       & B20--B25, B27--B36 & Emissive   \\
        \hline
    \end{tabular}
    \label{tab:modis-bands}
\end{table}

\begin{table}[h]
    \centering
    \caption{Overview of VIIRS bands and their resolutions. The bands are divided into reflective and emissive types.}
    \begin{tabular}{l|l|l}
        \multicolumn{3}{c}{\textbf{VIIRS Bands}}     \\
        \hline
        Resolution & Band               & Type       \\
        \hline
        375 m      & I1--I3             & Reflective \\
        375 m      & I4--I5             & Emissive   \\
        750 m      & M01--M06, M09      & Reflective \\
        750 m      & M07--M08, M10--M16 & Emissive   \\
        \hline
    \end{tabular}
    \label{tab:viirs-bands}
\end{table}

MODIS and VIIRS images have different projections that vary from day to day. For the prediction task, we want each data sample to relate to the same area from the current day to the next day. For this purpose, we need a common grid from which to retrieve samples. A simple geodetic grid was chosen, where each cell corresponds to 0.75 $\times$ 0.75 degrees longitude and latitude.
To avoid storing an excessive amount of data, the images (shown in Table \ref{tab:modis-bands} and \ref{tab:viirs-bands}) are resampled to a patch size that roughly matches their corresponding geodetic resolution at the equator. This corresponds to 64 pixels for 1 km data, 96 pixels for 750 m, and 192 pixels for 375 m. The geolocation data (Table \ref{tab:data-products}) provides geodetic coordinates for each pixel, which are converted to floating-point coordinates within each patch. Integer pixel values are avoided at this stage to preserve information for the final resampling step. We use inverse distance resampling for the raw sensor bands and nearest neighbor resampling for the fire mask data. The weather and drought data were already in a geodetic projection. Therefore, we simply patchify them and insert each patch into the corresponding cell (in the geodetic grid).

\subsection{Experiments}
The focus of the predictability study is wildfire spread, and not wildfire ignition. Therefore, we only select cells and dates where there is at least one fire pixel in both the MODIS and VIIRS data for the current day. The requirement that both datasets have a fire pixel ensures that the two models are trained on the same set of samples.

The input data is upsampled into a shared resolution corresponding to 192 $\times$ 192 pixels per patch for both MODIS and VIIRS models. This ensures that the VIIRS model can benefit from its higher resolution, while still allowing the MODIS model to potentially benefit from increased model capacity through upsampling. Both MODIS and VIIRS sensors capture data during both day and night, with only emissive bands included at night. The input features are then concatenated along the channel dimension. This results in 65 input channels for the MODIS model and 43 input channels for the VIIRS model.

The architecture used for both models is a U-Net \cite{unet} with a ResNet-50 backbone \cite{resnet}. This choice was made to keep the architecture simple and use proven technologies widely used in the field. The output is of the same spatial dimensions as the input, 192 $\times$ 192 pixels. However, to allow a fair comparison between the two fire products, the target fire masks are downsampled to the resolution of the MOD14 data (64 pixels). The loss and metrics are, therefore, calculated on the 64 $\times$ 64 resolution. The target fire mask is a max aggregation of the daytime and nighttime fire mask for the next day.
The loss function is binary cross entropy with a positive weight $w$ to account for the imbalance in the labels:
\begin{equation}
    L = -\frac{1}{N}\sum_{i=1}^N \big( w y_i \log(\hat{y}_i) + (1-y_i)\log(1-\hat{y}_i) \big) ,
\end{equation}
where $y_i$ is the true label, $\hat{y}_i$ is the predicted probability, and $N$ is the number of pixels.
The positive weight, $w = 3$, was found through an ablation study. The testing is done with the checkpoint that achieves the highest intersection over union (IoU) on the validation set.
The IoU is performed pixel-wise across all samples.

To provide further basis for the suitability of one fire product for the prediction task, we also include a qualitative assessment of the MOD14 and VNP14 fire masks. The purpose of this exercise is to reason about why one of the datasets might perform better than the other for the prediction task. To do this, we pick some of the larger fires that occurred in the dataset. Then, for each of the fires, we plot the progression of the fire according to both VNP14 and MOD14. We then reason about the relative realism of the fire progression of MOD14 versus VNP14.

\section{Results and discussion}
\begin{table}[htbp]
    \centering
    \caption{Evaluation of the persistence baselines on the test set.}
    \begin{tabular}{l|cc}
        \hline
        Product & F1 Score (\%) & IoU (\%) \\
        \hline
        MOD14   & 6.56          & 3.39     \\
        VNP14   & 19.62         & 10.87    \\
        \hline
    \end{tabular}
    \label{tab:temporal_consistency}
\end{table}

\begin{table}[htbp]
    \centering
    \caption{Test results for the two models trained exclusively on either MODIS/MOD14 or VIIRS/VNP14 data. VIIRS/VNP14 is clearly superior.}
    \begin{tabular}{lll|cc}
        \hline
        Input & \makecell{Training                                       \\Target} & \makecell{Evaluation\\Target} & \makecell{F1 Score\\(\%)} & IoU (\%) \\
        \hline
        MODIS & MOD14              & MOD14 & 11.96 ± 1.52 & 6.37 ± 0.86  \\
        VIIRS & VNP14              & VNP14 & 28.57 ± 0.62 & 16.67 ± 0.42 \\
        \hline
    \end{tabular}
    \label{tab:model_comparison1}
\end{table}

\begin{figure}[htbp]
    \centering
    \begin{tabular}{c}
        \includegraphics[width=0.40\textwidth,trim=30 70 30 70,clip]{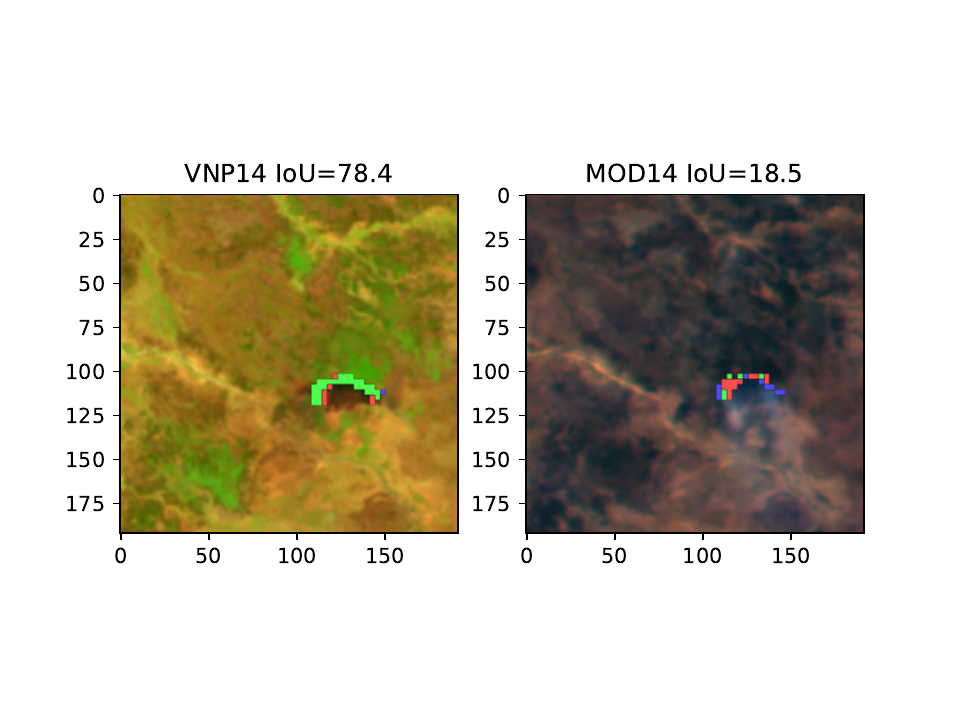} \\
        \includegraphics[width=0.40\textwidth,trim=30 70 30 70,clip]{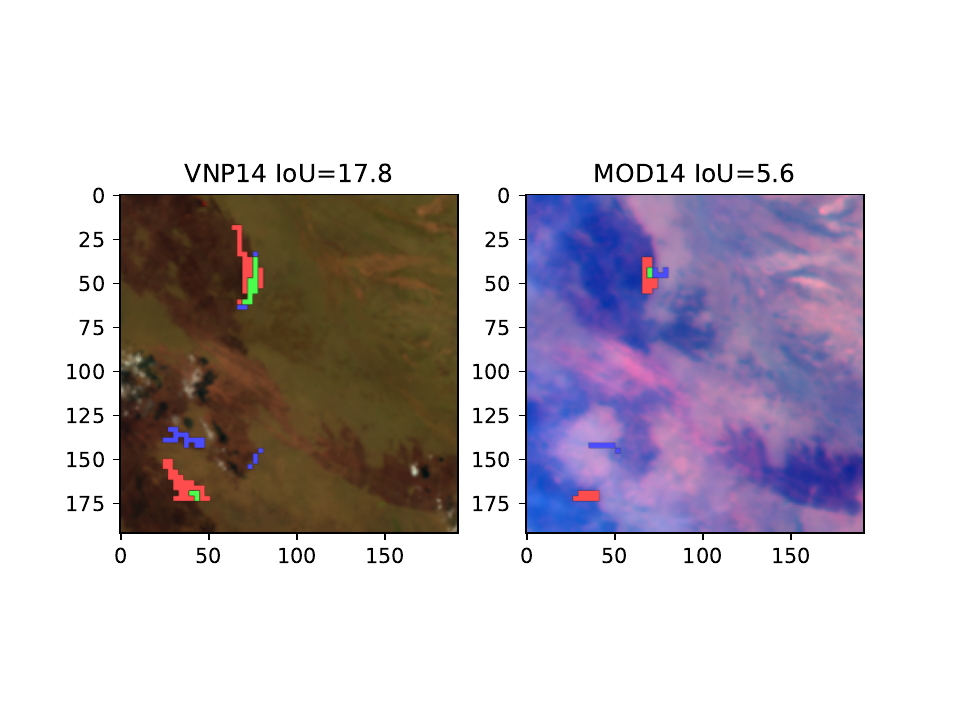} \\
    \end{tabular}
    \caption{Sample predictions from the two models in Table \ref{tab:model_comparison1}. Areas filled in green indicate \colorbox{green!30}{true positives} (correctly predicted fires), red indicates \colorbox{red!30}{false positives} (incorrectly predicted fires), and blue indicates \colorbox{blue!30}{false negatives} (missed fires).}
    \label{fig:samples}
\end{figure}

\begin{table}[htbp]
    \centering
    \caption{Test results where we swap the input used for each model. We can see that the problem lies in the MOD14 fire masks, and not the MODIS data.}
    \begin{tabular}{lll|cc}
        \hline
        \makecell{Input} & \makecell{Training                                       \\Target} & \makecell{Evaluation\\Target} & \makecell{F1 Score\\(\%)} & \makecell{IoU\\(\%)} \\
        \hline
        MODIS            & VNP14              & VNP14 & 27.16 ± 0.63 & 15.72 ± 0.42 \\
        VIIRS            & MOD14              & MOD14 & 16.09 ± 0.59 & 8.75 ± 0.35  \\
        \hline
    \end{tabular}
    \label{tab:model_comparison2}
\end{table}

\begin{table}[htbp]
    \centering
    \caption{Measuring the impact on learning by swapping training and testing data sources.}
    \begin{tabular}{lll|cc}
        \hline
        Input & \makecell{Training                                      \\Target} & \makecell{Evaluation\\Target} & \makecell{F1 Score\\(\%)} & IoU (\%) \\
        \hline
        MODIS & MOD14              & VNP14 & 17.67 ± 2.81 & 9.72 ± 1.70 \\
        MODIS & VNP14              & MOD14 & 15.11 ± 0.91 & 8.17 ± 0.53 \\
        VIIRS & VNP14              & MOD14 & 17.03 ± 1.18 & 9.31 ± 0.71 \\
        VIIRS & MOD14              & VNP14 & 16.66 ± 3.13 & 9.12 ± 1.86 \\
        \hline
    \end{tabular}
    \label{tab:model_comparison3}
\end{table}
\begin{figure}[htbp]
    \centering
    \begin{tabular}{cc}
        \includegraphics[width=0.45\textwidth,trim=90 80 40 40,clip]{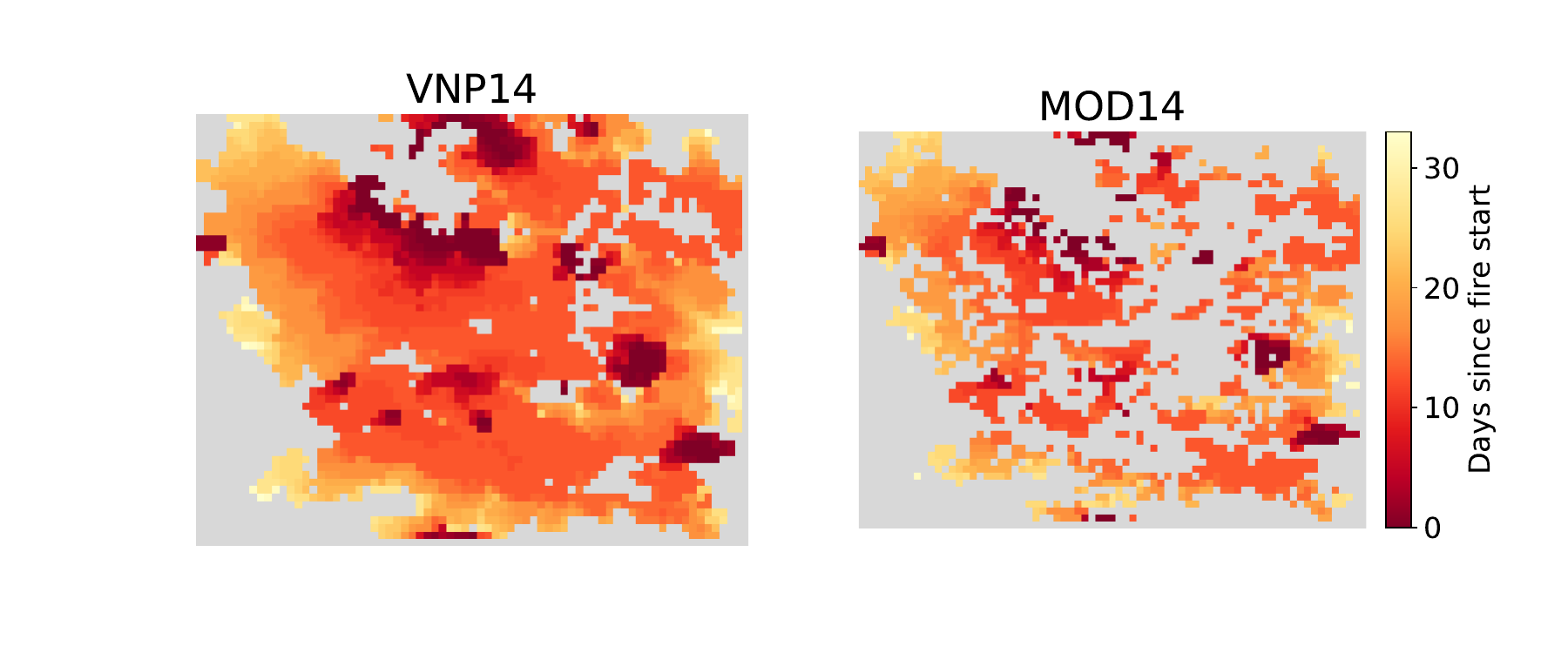} \\
        \includegraphics[width=0.45\textwidth,trim=90 80 40 40,clip]{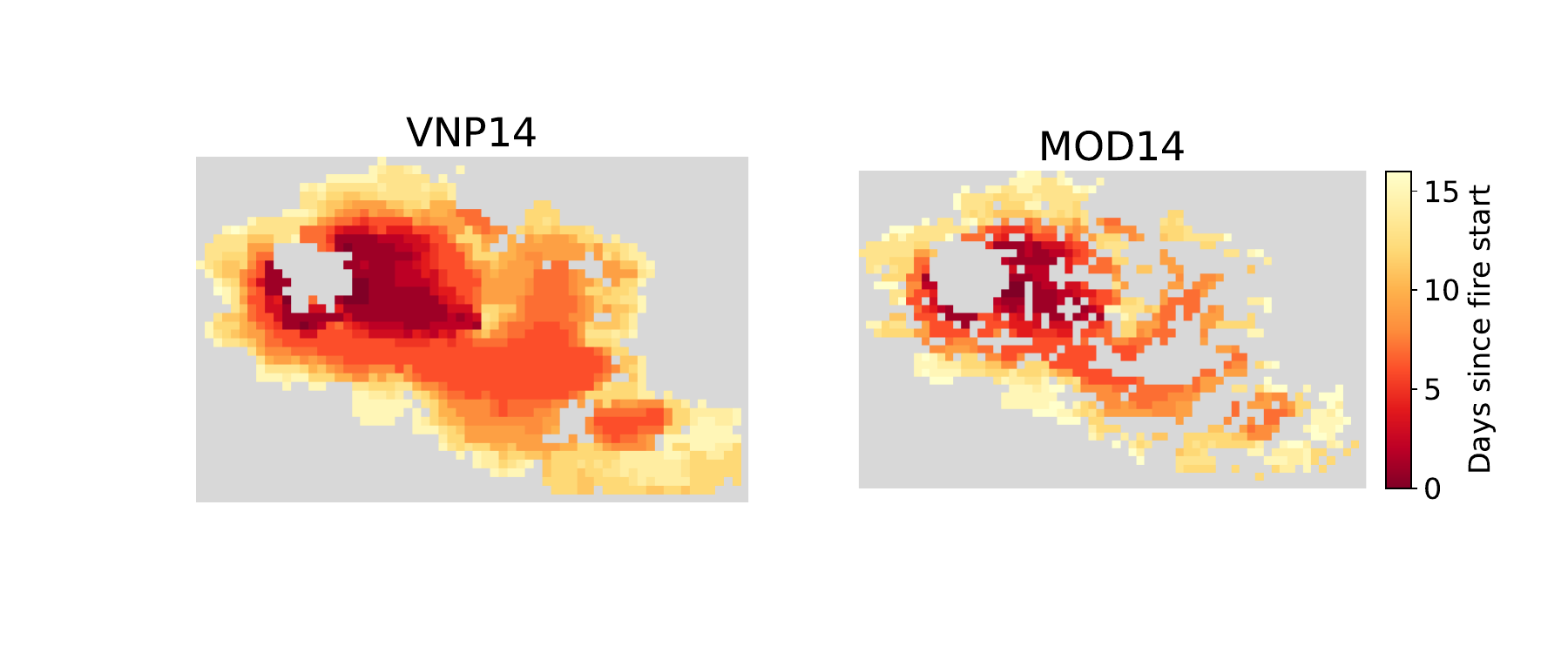} \\
    \end{tabular}
    \caption{Qualitative assessment of the wildfire spread for both products. The color scale shows progression of each individual fire measured from the ignition date of said fire.}
    \label{fig:qualitative_assesment}
\end{figure}

We first present results for the persistence baselines, representing a lower bound for the expected performance of the models. We then present results for the two models trained exclusively on either MODIS+MOD14 or VIIRS+VNP14 data. To investigate the underlying reasons for the performance difference, we also present results of models that use a mix of MODIS, MOD14, VIIRS, and VNP14 data for the three different data slots in the experiment: input, training, and evaluation. To increase the confidence of the results, 5 independent runs are done and all metrics are presented with their means and standard deviations.
We then compare the fire progression between MOD14 and VNP14 for a select number of individual fires and find large differences in the apparent quality of the fire masks. Finally, we discuss the results and possible implications for the use of MODIS and VIIRS data for wildfire prediction.

\subsection{Predictability}

We establish a \emph{persistence baseline} for the prediction task by measuring the performance of assuming the next-day fires will equal the current day fires. The results for this baseline are reported in Table \ref{tab:temporal_consistency} for both MOD14 and VNP14. This baseline is also a measurement of the temporal consistency of the products. We can clearly see that VNP14 is more consistent than MOD14.

We train two models that are fed with two different data sources: MODIS (with MOD14 fire masks) and VIIRS (with VNP14 fire masks). The results are shown in Table \ref{tab:model_comparison1}.

In Figure \ref{fig:samples}, we can see some sample predictions for the two models. Both of these models beat their respective persistence baselines established in Table \ref{tab:temporal_consistency} by a large margin. We can see that the model trained with VIIRS data performs much better than the MODIS model. There could be multiple reasons for this:
\begin{enumerate}
    \item The high-resolution VIIRS raw band data as input to the model gives superior prediction performance.
    \item The VNP14 fire masks are more consistent and easier to predict.
\end{enumerate}

We, therefore, also train two models where input (MODIS or VIIRS bands) and target (MOD14 or VNP14 fire masks) are swapped. The results are shown in Table \ref{tab:model_comparison2}. We can see that the main problem with the MODIS model was in fact the MOD14 fire masks, and not the MODIS input data. Furthermore, predicting VNP14 from MODIS data is a harder task than the other way around, since VIIRS captures data approximately 2 hours later. Using VIIRS input data helps with predicting both VNP14 and MOD14, although the increase in performance is not as significant.

We have thus narrowed down that the biggest detractor from performance is the MOD14 fire masks. Finally, we want to evaluate how much MOD14 as a training target hinders learning. We perform this comparison by training four additional models where the training and evaluation targets are different. The results are shown in Table \ref{tab:model_comparison3}. If we compare row 1 in Table \ref{tab:model_comparison1} with row 2 in Table \ref{tab:model_comparison3}, we can see that the performance is lower when training on MOD14 instead of VNP14, \emph{even when testing} on MOD14 data. By training on VNP14 instead of MOD14, we get a mean IoU of 8.17\% when testing on MOD14, compared with 6.37\% when training \emph{and testing} on MOD14 data. Put another way, for the machine learning model evaluated, VNP14 is a better prior for new MOD14 data than MOD14 data itself.
\subsection{Qualitative assessment}
In Figure \ref{fig:qualitative_assesment}, we can see a comparison between the progression of VNP14 and MOD14 fires. A higher-level comparison can be seen in Figure \ref{fig:south-east-australia}. It is clear that VNP14 produces a more cohesive and uniform fire spread. MOD14 seems to contain stochastic detections of fire. It is reasonable to assume that the MOD14 models struggle with this stochastic nature of detections. The input data does not support why the fire spreads in seemingly random and erratic patterns, and so the learned patterns will not generalize to the test set.

\subsection{Discussion}
The results in Table \ref{tab:model_comparison2} show that the problem does not lie in the quality of the raw bands. When using MODIS input bands but VIIRS target, the model performs much better than the other way around. This supports the conclusion that MOD14 provides an unreliable target particularly ill-suited for the task of next-day prediction using machine learning. A further indicator of this is the qualitative results in Figure \ref{fig:qualitative_assesment}. The target data forces the model to learn irrational patterns, and the model is unable to generalize well to the test set. While stochastic detections might not be detrimental for all applications, they are a poor match for this particular task.

As stated previously, MOD14 and VNP14 rely on the same algorithmic basis. Why do we then see such a difference in performance? There is a significant difference in spatial resolution for the infrared bands used to detect temperature anomalies. VNP14 has a spatial resolution of 375 m, while MOD14 has a resolution of 1 km.
However, as we saw in Table \ref{tab:model_comparison2}, we were able to achieve good results using MODIS input and VNP14 target. The increased capability of a deep learning model might be able to compensate for the lower resolution of the MODIS data.

\section{Conclusion}

In this study, we systematically compare the predictability of MOD14 and VNP14 for next-day fire prediction using deep learning models. Our experiments show that the MOD14 fire mask is highly unpredictable, while VNP14 proves to be better suited for this task. The unreliability of MOD14 is not explained by the quality of the raw bands of MODIS, since using them as input and trying to predict VIIRS had much better results. The relative success of using MODIS input and VNP14 target indicates that an improved fire detection model using deep learning is possible for MODIS, which we plan to explore in future work.

\section*{Acknowledgment}
This work was supported in part by the Excellence Center at Linköping-Lund in Information Technology (ELLIIT) Researcher Funding, the Zenith Research Program, the Swedish Research Council with grant agreement no. 2016-05543 and no. 2024-05652, and the Wallenberg Artificial Intelligence, Autonomous Systems and Software Program (WASP), funded by Knut and Alice Wallenberg Foundation. The computational resources were provided by the National Academic Infrastructure for Supercomputing in Sweden (NAISS) at C3SE, and by the Berzelius resource, provided by the Knut and Alice Wallenberg Foundation at the National Supercomputer Centre.

\bibliographystyle{splncs04}
\bibliography{main}

\end{document}